\documentclass[10pt,twocolumn,letterpaper]{article}

\usepackage[pagenumbers]{cvpr} % To force page numbers, e.g. for an arXiv version

\definecolor{cvprblue}{rgb}{0.21,0.49,0.74}
\usepackage[pagebackref,breaklinks,colorlinks,allcolors=cvprblue]{hyperref}

\usepackage[table,xcdraw]{xcolor}
\usepackage{multirow}
\usepackage{placeins}

\def\confName{CVPR}
\def\confYear{2026}

\title{\includegraphics[height=1em, keepaspectratio]{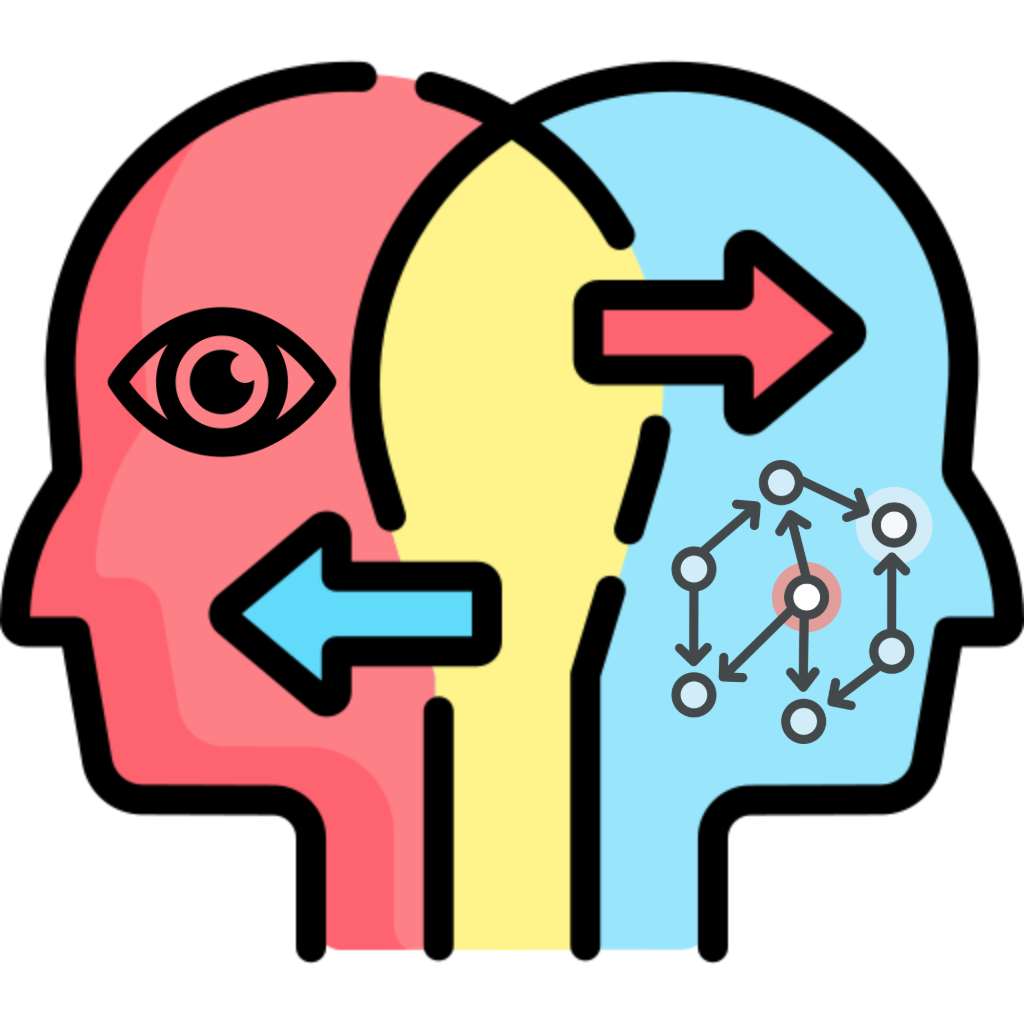}~CLiViS: \\Unleashing Cognitive Map through Linguistic-Visual Synergy for Embodied Visual Reasoning}

\author{%
\begin{tabular}{c@{\hspace{1em}}c@{\hspace{1em}}c@{\hspace{1em}}c@{\hspace{1em}}c@{\hspace{1em}}c}
    Kailing Li\textsuperscript{\rm 1}\footnotemark[1]&Qi'ao Xu\textsuperscript{\rm1}\footnotemark[1]&Tianwen Qian\textsuperscript{\rm1}\footnotemark[1] \kern0.2em\footnotemark[2] &Yuqian Fu\textsuperscript{\rm2}&Yang Jiao\textsuperscript{\rm3}&Xiaoling Wang\textsuperscript{\rm1}\footnotemark[2] 
  \end{tabular}\\
  \\
  \textsuperscript{1}School of Computer Science and Technology, East China Normal University \\
  \textsuperscript{2}King Abdullah University of Science and Technology, 
  \textsuperscript{3}Fudan University\\
\small{\texttt{51275901046@stu.ecnu.edu.cn, twqian@cs.ecnu.edu.cn}}
}

\begin{document}

% 在 maketitle 之前把脚注改成符号样式 (1 变 *, 2 变 dagger)
\renewcommand{\thefootnote}{\fnsymbol{footnote}}

\maketitle

\footnotetext[1]{Equal contribution.}
\footnotetext[2]{Corresponding author.}

% 如果正文里还需要用正常的 1, 2, 3 脚注，记得切回来
\renewcommand{\thefootnote}{\arabic{footnote}}

\maketitle
\begin{abstract}
Embodied Visual Reasoning (EVR) seeks to follow complex, free-form instructions based on egocentric video, enabling semantic understanding and spatiotemporal reasoning in dynamic environments. Despite its promising potential, EVR encounters significant challenges stemming from the diversity of complex instructions and the intricate spatiotemporal dynamics in long-term egocentric videos. Prior solutions either employ Large Language Models (LLMs) over static video captions, which often omit critical visual details, or rely on end‑to‑end Vision-Language Models (VLMs) that struggle with stepwise compositional reasoning. Considering the complementary strengths of LLMs in reasoning and VLMs in perception, we propose \textbf{CLiViS}. It is a novel training-free framework that leverages LLMs for high-level task planning and orchestrates VLM‑driven open‑world visual perception to iteratively update the scene context. Building on this synergy, the core of CLiViS is a dynamic \textbf{Cognitive Map} that evolves throughout the reasoning process. This map constructs a structured representation of the embodied scene, bridging low-level perception and high-level reasoning. Extensive experiments across multiple benchmarks demonstrate the effectiveness and generality of CLiViS, especially in handling long‑term visual dependencies. Code is available at \url{https://github.com/Teacher-Tom/CLiViS}.
\end{abstract}    
\section{Introduction}
\label{sec:intro}

Embodied Visual Reasoning (EVR), also known as Embodied Question Answering (EQA) \cite{das2018embodied}, aims to follow free-form user instructions based on egocentric video input, enabling semantic understanding and spatiotemporal reasoning in dynamic environments. With its broad application potential in fields such as robotics \cite{ma2024survey, he2025sequential, wang2026ocra} and autonomous driving \cite{grigorescu2020survey, brodermann2025cafuser}, EVR has attracted significant research interest in recent years.

Despite its potential, EVR faces two key challenges: First, egocentric video is characterized by long temporal sequences and a limited field of view, leading to difficulty in spatiotemporal perception. Second, user instructions exhibit complexity and diversity, imposing significant demands on compositional reasoning. To address these challenges, EVR models must not only support accurate open-vocabulary visual perception, but also perform compositional reasoning by grounding complex instructions into a sequence of vision-language subtasks, such as key object localization, recognition, and captioning, and integrating them coherently to support the final decision.

\begin{figure}
    \centering
    \includegraphics[width=0.9\linewidth]{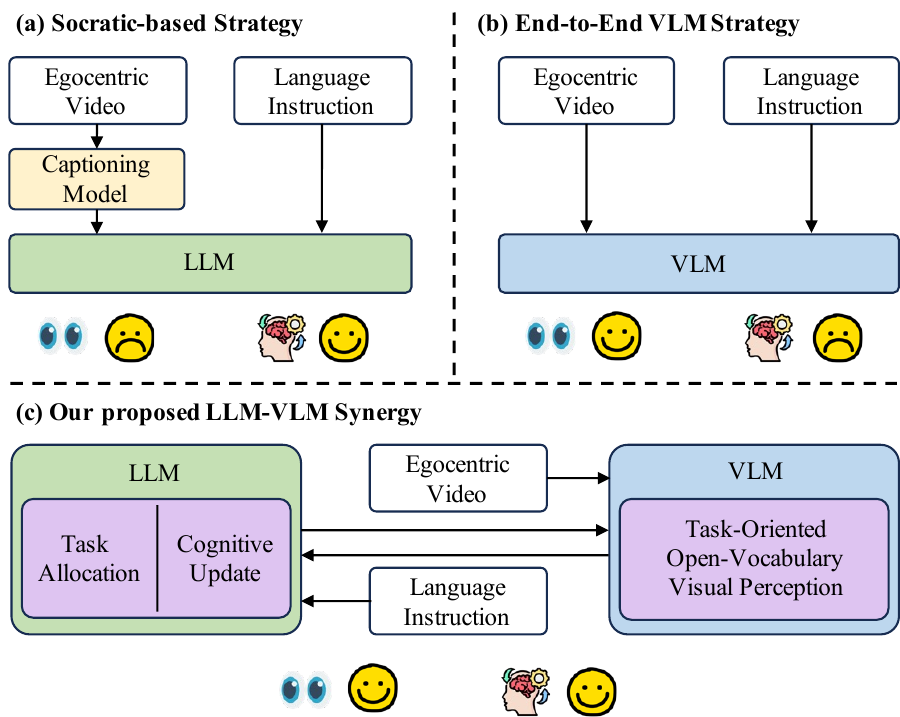}
    \caption{\textbf{Comparison of our proposed CLiViS with previous methods.} CLiViS bridges perception and reasoning by combining the strengths of LLMs and VLMs.}
    \label{fig:teaser}
\end{figure}

As illustrated in Figure \ref{fig:teaser} (a) and (b), existing methods can be broadly divided into two categories, but neither of them can balance the challenges of perception and reasoning. \textbf{Socratic-based strategy} \cite{kim2023context, zeng2023socratic} first adopts a video captioning model to translate video content into text descriptions, which are then combined with instructions and processed by Large Language Models (LLMs) for reasoning. This design fully leverages the advanced semantic reasoning capabilities of LLMs, but inevitably suffers from fixed, instruction‑agnostic captions that omit fine‑grained, context‑sensitive details. In contrast, another line of work relies on \textbf{End-to-End VLM} to integrate visual perception and reasoning within a unified framework \cite{danny2023Palme, gupta2024open}. By fusing vision and language modalities at the feature level, these approaches mitigate the information loss inherent in cross-modal conversion, and exhibit potent open-vocabulary perception capabilities that can flexibly adapt to diverse instructions. However, current VLMs remain deficient in high-level logical planning and semantic reasoning, affecting their ability to understand and execute complex instructions. Consequently, these approaches struggle to organize essential reasoning steps in a coherent and task-driven manner, such as event localization, object recognition, and relation extraction, thus failing to fully exploit their powerful perceptual strengths.
To alleviate this limitation, recent efforts have explored \textbf{Video Reasoning}. For example, Video-R1 \cite{feng2025video} employs rule-based reinforcement learning to activate the reasoning capabilities of Video-LLM. VideoAgent \cite{fan2024videoagent} and VideoTree \cite{wang2024videotree} enrich the visual context by dynamically sampling frames under LLM guidance. However, these methods either incur substantial training costs or reduce the LLM to a passive selector, failing to fully leverage its potential for high-level planning and structured reasoning.

Building on these insights, we propose \textit{unleashing \textbf{C}ognitive map through \textbf{Li}nguistic-\textbf{Vi}sual \textbf{S}ynergy} (\textit{\textbf{CLiViS}}), a novel training-free framework that leverages the complementary strengths of pretrained LLMs and VLMs. As shown in Figure \ref{fig:teaser} (c), in our approach, the LLM acts as a high-level planner that assigns a sequence of subtasks to the VLM based on the instruction and accumulated scene cognition. These subtasks range from key object recognition to dense object captioning and relation extraction, \textit{etc}. Each subtask guides the VLM to extract task-relevant visual cues to update the scene cognition, which are then integrated to infer the final response by LLM. The core of CLiViS is a dynamic \textit{\textbf{Cognitive Map}}, which is continuously refined through the interaction between the LLM and VLM. This map offers a structured representation of egocentric videos, encompassing region and event segmentation, spatiotemporal details of key objects, agents, actions, and their semantic relationships. It consists of two components: a scene navigation graph and an object relation graph. The former divides the video into multiple segments at equal intervals and records the areas, entities, actions, and their captions involved in each segment. The latter meticulously records dense object attributes and their semantic interactions, providing a rich context for complex reasoning. Guided by the subtasks assigned by the LLM, the VLM extracts relevant visual cues and incorporates them into the evolving cognitive map, effectively bridging fine-grained visual perception with high-level semantic reasoning. To further enhance the interpretability of the reasoning, we introduce an evidence memory module that accumulates instruction-relevant rationales extracted by VLM at each reasoning step. Finally, the cognitive map and evidence memory are integrated to facilitate the final response.

In summary, our contributions are as follows:
\begin{itemize}
    \item We introduce a novel training-free framework CLiViS, which synergizes LLMs and VLMs to tackle the dual challenges of perception and reasoning in embodied visual reasoning.
    \item We design a dynamic cognitive map that provides a structured semantic representation of the embodied scene, which evolves throughout the reasoning process and serves as a bridge between the LLM and VLM collaboration.
    \item We conduct extensive experiments across multiple egocentric benchmarks and foundation models, demonstrating the effectiveness and generalizability of CLiViS.
\end{itemize}

\section{Related Works}
\label{sec:rel}

\textbf{Embodied Question Answering.}
Embodied Question Answering (EQA) requires agents to integrate vision, language, and robotics to perceive and act within first-person environments based on language instructions \cite{liu2024aligning}. EQA is typically categorized into Active EQA (A-EQA), which involves interactive navigation, and Episodic-Memory EQA (EM-EQA), which reasons over pre-collected offline videos \cite{majumdar2024openeqa}. EM-EQA's primary challenge lies in efficiently extracting and integrating spatio-temporal information from passively captured visual signals. Due to data availability constraints, our work concentrates on the EM-EQA setting. Prominent benchmarks for this task include OpenEQA \cite{majumdar2024openeqa}, EgoTaskQA \cite{jia2022egotaskqa}, and EgoSchema \cite{mangalam2023egoschema}. Methodologically, existing approaches generally follow two strategies: (1) converting visual inputs into text for LLM reasoning, or (2) encoding video frames via vision models into LLM latent spaces. For instance, some methods employ tracking or segmentation models to capture inter-frame object correspondences \cite{yan2021lighttrack, fu2025objectrelator, pan2025v}, while others leverage foundation models and parameter-efficient fine-tuning on VLMs \cite{suglia2024alanavlm}. In contrast, our method synergistically combines the strengths of both VLMs and LLMs to construct an EQA framework without additional training.

\noindent
\textbf{LLM and VLM for Video Understanding.}
Large Language Models (LLMs) \cite{radford2018improving, touvron2023llama} provide powerful language understanding and reasoning. This foundation was extended by Multimodal LLMs (MLLMs) like LLaVA \cite{liu2023visual} and Qwen-VL \cite{Qwen-VL}, which map visual features into the LLM's latent space to support image-text tasks. However, video presents greater complexity due to its dynamic spatio-temporal nature, challenging standard MLLMs. Consequently, Video-LLMs emerged, integrating temporal encoders \cite{zhang2023video} and specialized memory mechanisms or training strategies \cite{song2024moviechat, huang2024vtimellm} to capture long-range dependencies and action dynamics. Despite progress in general video understanding, significant challenges remain in embodied tasks. The first-person perspective, characterized by frequent subject–object interactions and a limited field-of-view, further complicates the spatio-temporal reasoning required for these tasks.

\noindent
\textbf{Episodic Memory for LLM Agents.}
LLM agents leverage episodic memory to integrate long-term interactions \cite{pink2025position}. The format of memory determines how information is stored, managed, and utilized. Existing works mainly divide into two forms: parameter-based \cite{shao2023character, tiwari2024mac} and text-based \cite{packer2023memgpt, hu2023chatdb}. Parameter-based approaches implicitly encode memory within model parameters, and update it via fine-tuning, \textit{e.g.}, Character-LLM's role-specific fine-tuning \cite{shao2023character} and MAC’s online weight adaptation \cite{tiwari2024mac}. Text-based approaches explicitly store memory in natural language or structured symbols, and provide it as prompts to LLM without parameter updating, \textit{e.g.}, MemGPT’s \cite{packer2023memgpt} working contexts, ChatDB’s \cite{hu2023chatdb} SQL-retrievable symbolic memory, and ReMEmbR's~\cite{anwar2025remembr} vector database memory.

\section{Method}

\begin{figure*}
    \centering
    \includegraphics[width=\linewidth]{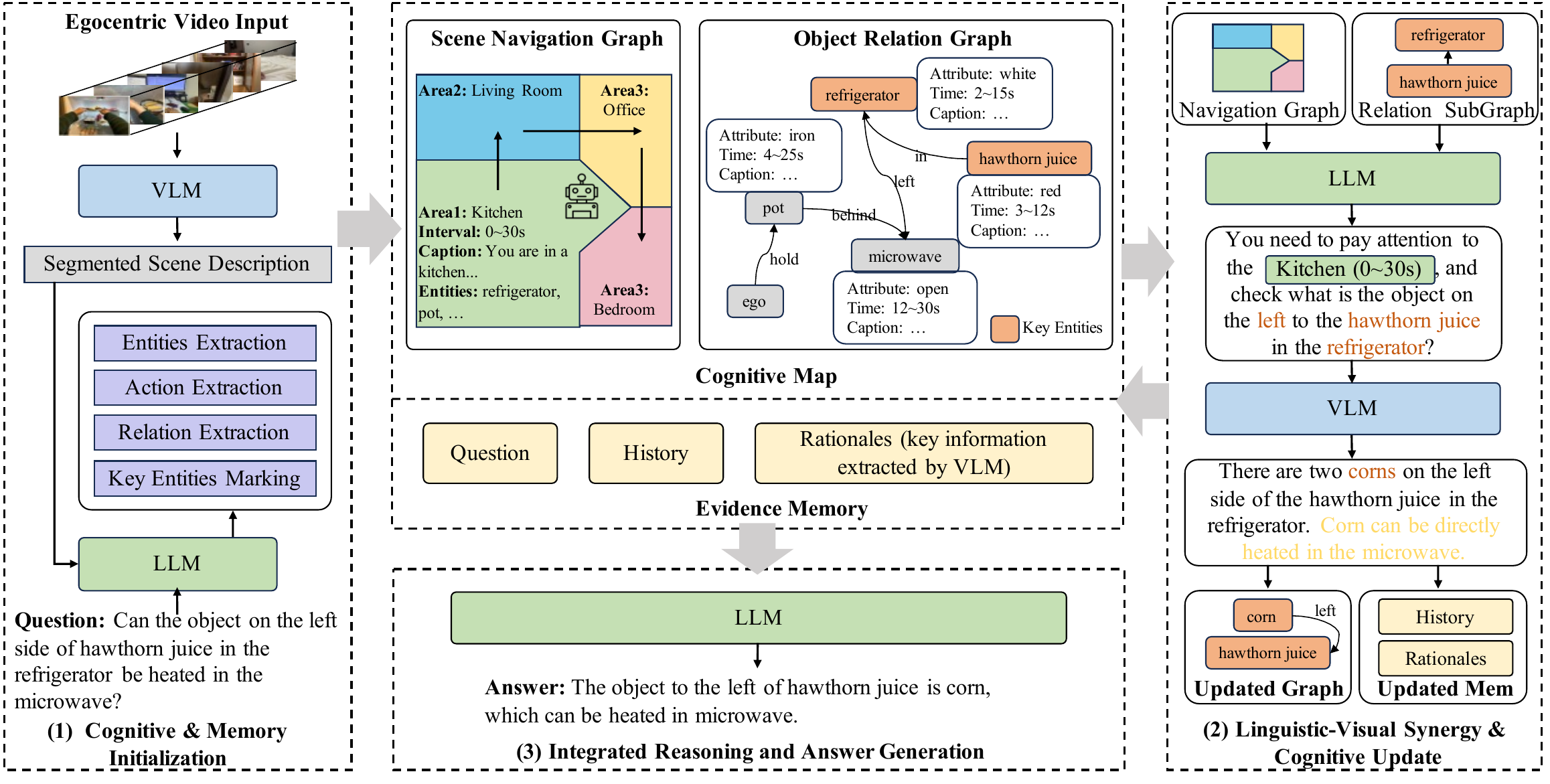}
    \caption{\textbf{Overall Framework.} The inference of CLiViS consists of three steps: (1) initialize the cognitive map and evidence memory from segmented scene descriptions; (2) iteratively update task-relevant visual cues via LLM–VLM interaction; and (3) integrate context for final answer generation.}
    \label{fig:framework}
\end{figure*}

We propose \textbf{CLiViS}, a novel framework for embodied visual reasoning that constructs dynamic cognitive map through LLM–VLM collaboration. CLiViS combines the advanced reasoning capabilities of LLMs with the open-vocabulary perception of VLMs to model embodied scenes. In Sec. \ref{ssec:formulation}, we provide a formal definition of the EVR task, and redefine it based on our proposed modeling paradigm. Sec. \ref{ssec:framework} presents a framework overview of CLiViS. In Sec. \ref{ssec:cognitive_map}, we detail the structure of the cognitive map and evidence memory, as well as their update mechanism. Finally, in Sec. \ref{ssec:synergy}, we discuss the collaborative process between the LLM and VLM, with a focus on the design of prompts.

\subsection{Problem Formulation‌}
\label{ssec:formulation}

Embodied Visual Reasoning (EVR) is designed to achieve semantic understanding and spatiotemporal reasoning on dynamic scenes based on first-person videos and free-form language instructions, ultimately generating a response corresponding to the given instruction. Formally, the task can be defined as:
\begin{equation}
    R = f_\theta(V, I),
\end{equation}
where $V=\{v_1, v_2, ..., v_t\}$ represents the input video, $I$ and $R$ denote the instruction and the response respectively.

In the era of large-scale foundation models, EVR methods typically fall into two categories. LLM-based approaches follow a two-stage paradigm: a video captioning model first converts visual frames into textual descriptions, which are then concatenated with the instruction for answer reasoning via LLM:
\begin{equation}
    R = \text{LLM}(\text{Cap}(V), I),
\end{equation}
where $\text{Cap}$ represents the captioning model. In contrast, VLM-based approaches use the versatile vision-language understanding capabilities of VLM to perform end-to-end reasoning directly over visual inputs and instructions:
\begin{equation}
    R = \text{VLM}(V, I).
\end{equation}

To fully leverage the complementary strengths of LLMs and VLMs in reasoning and perception, we reformulate EVR as a synergy task that builds a dynamic cognitive map through LLM-VLM collaboration to support LLM reasoning:
\begin{equation}
    R = \text{LLM}\left( \mathcal{M}, I \mid \mathcal{M} = \bigcup_{T_i \in \text{LLM}(I, \mathcal{M})} \text{VLM}(V, T_i) \right),
\end{equation}
where $\mathcal{M}$ denotes the cognitive map, and $T = \left \{  T_i \right \} _{i=0}^{n-1}$ represents a series of sub-tasks decomposed by the LLM based on the known information $I$ and $\mathcal{M}$. We elaborate on these concepts and the associated workflow in the following sections.

\subsection{CLiViS Framework Overview‌}
\label{ssec:framework}

The overall framework of CLiViS is illustrated as Figure \ref{fig:framework}. The inference process unfolds through three stages, enabling synergistic collaboration between LLM and VLM for embodied visual reasoning.
\textbf{(1) Cognitive and Memory Initialization:} Given an egocentric video $V$ and the corresponding language instruction $I$, we begin by segmenting the video into fixed-length clips (\textit{e.g.}, 30s). For each segment, the VLM is invoked to generate coarse-grained visual descriptions. These descriptions are then analyzed by the LLM, which extracts salient entities (\textit{e.g.}, objects, regions, persons), actions, and inter-entity relationships. Conditioned on the instruction $I$, the LLM further identifies and highlights the entities that are most relevant to the question. Based on this information, an initial cognitive map is constructed, encompassing entities (\textit{e.g.},  ``refrigerator'') and their attributes (\textit{e.g.}, ``white''), actions (\textit{e.g.}, ``hold''), and relations (\textit{e.g.}, ``inside''). Simultaneously, an evidence memory module is initialized to store the question, interaction history, and task-relevant rationales. \textbf{(2) Linguistic-Visual Synergy and Cognitive Update:} In an iterative reasoning loop, the LLM integrates current cognitive map and evidence memory to assess whether sufficient information has been gathered to answer the question. If not, it will generate a sub-instruction (\textit{e.g.}, ``check what is on the left of the hawthorn juice in the refrigerator'') for a specific temporal segmentation (\textit{e.g.}, ``kitchen (0\textasciitilde30s)'') to guide the VLM in performing focused perception. Subsequently, the responses from the VLM will be parsed to extract key entities, relations, and supporting rationales, which will then be used to update both the cognitive map and evidence memory. \textbf{(3) Integrated Reasoning and Answer Generation:} Once the LLM determines that sufficient information has been collected to answer the question, or the reasoning process reaches the predefined maximum number of iterations, it will generate the final answer. This design ensures a tight integration between perception and reasoning.

\begin{figure*}
    \centering
    \includegraphics[width=\linewidth]{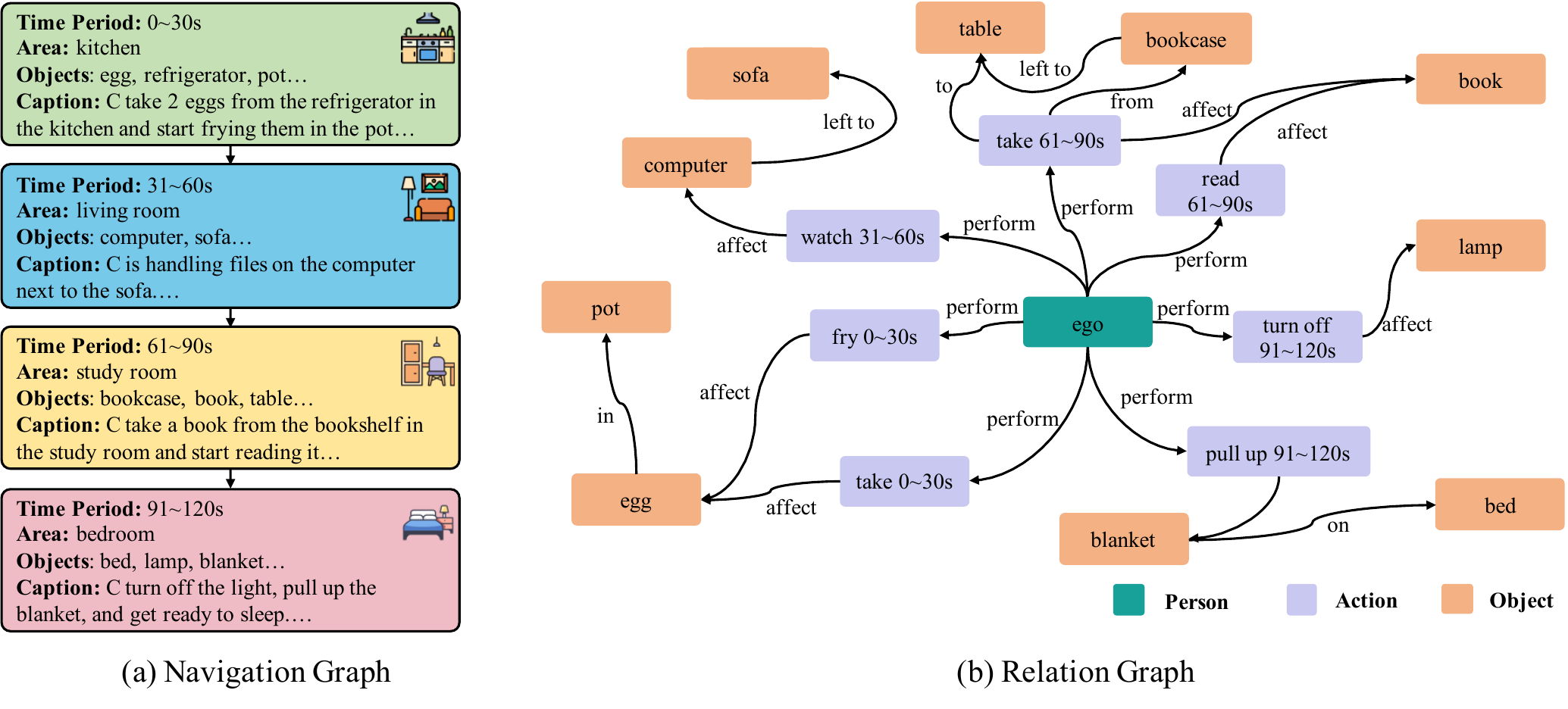}
    \caption{\textbf{Cognitive Map consists of a navigation graph and a relation graph.} The former captures temporal regions and associated entities. The latter records fine-grained relations between entities.}
    \label{fig:graph}
\end{figure*}

\subsection{Cognitive Map and Evidence Memory}
\label{ssec:cognitive_map}

\textbf{Cognitive Map Definition.}
To enable structured reasoning in embodied scenarios, we define the cognitive map $\mathcal{M}$ as a graph-based representation of key regions, entities, actions, and their relationships, built through linguistic–visual interactions. Formally, it consists of two subgraphs:
\begin{equation}
    \mathcal{M}=\left \{ \mathcal{G}_{nav}, \mathcal{G}_{rel} \right \} = \left \{ \left ( V_{nav}, E_{nav} \right ), \left ( V_{rel}, E_{rel} \right )  \right \},
\end{equation}
where $\mathcal{G}_{nav}$ is the \textbf{Navigation Graph} (Figure \ref{fig:graph} (a)), capturing the temporal structure of the video. Each node $v_i \in V_{nav}$ denotes a distinct time segment, and edges $e_{ij} \in E_{nav}$ indicate temporal adjacency between segments.
$\mathcal{G}_{rel}=\left ( V_{rel}, E_{rel} \right ) $ is the \textbf{Relation Graph} (Figure \ref{fig:graph} (b)), modeling fine-grained entity-level relationships. Each node $v_i \in V_{rel}$ represents a visual entity or action, and edges $e_{ij} \in E_{rel}$ denote semantic relations between the nodes, such as spatial relationship, agent-object interaction, and functional dependency.

\noindent
\textbf{Evidence Memory Definition.}
To complement the perception-centric cognitive map, we introduce an Evidence Memory $\mathcal{E}$ to retain high-level semantic cues derived from LLM–VLM interactions. $\mathcal{E}$ serves as a lightweight buffer that accumulates reasoning-relevant evidence tied to the instruction. Specifically, LLM prompts VLM to analyze the corresponding video clips, and distills the VLM's response into an evidence atom. Formally, the evidence atom is defined as:
\begin{equation}
    \mathcal{E} = (r, \tau, O),
\end{equation}
where $r$ is a language rationale about the query, $\tau$ represents the associated time span. $O$ denotes the set of objects, regions or actions involved.

\begin{figure*}
    \centering
    \includegraphics[width=\linewidth]{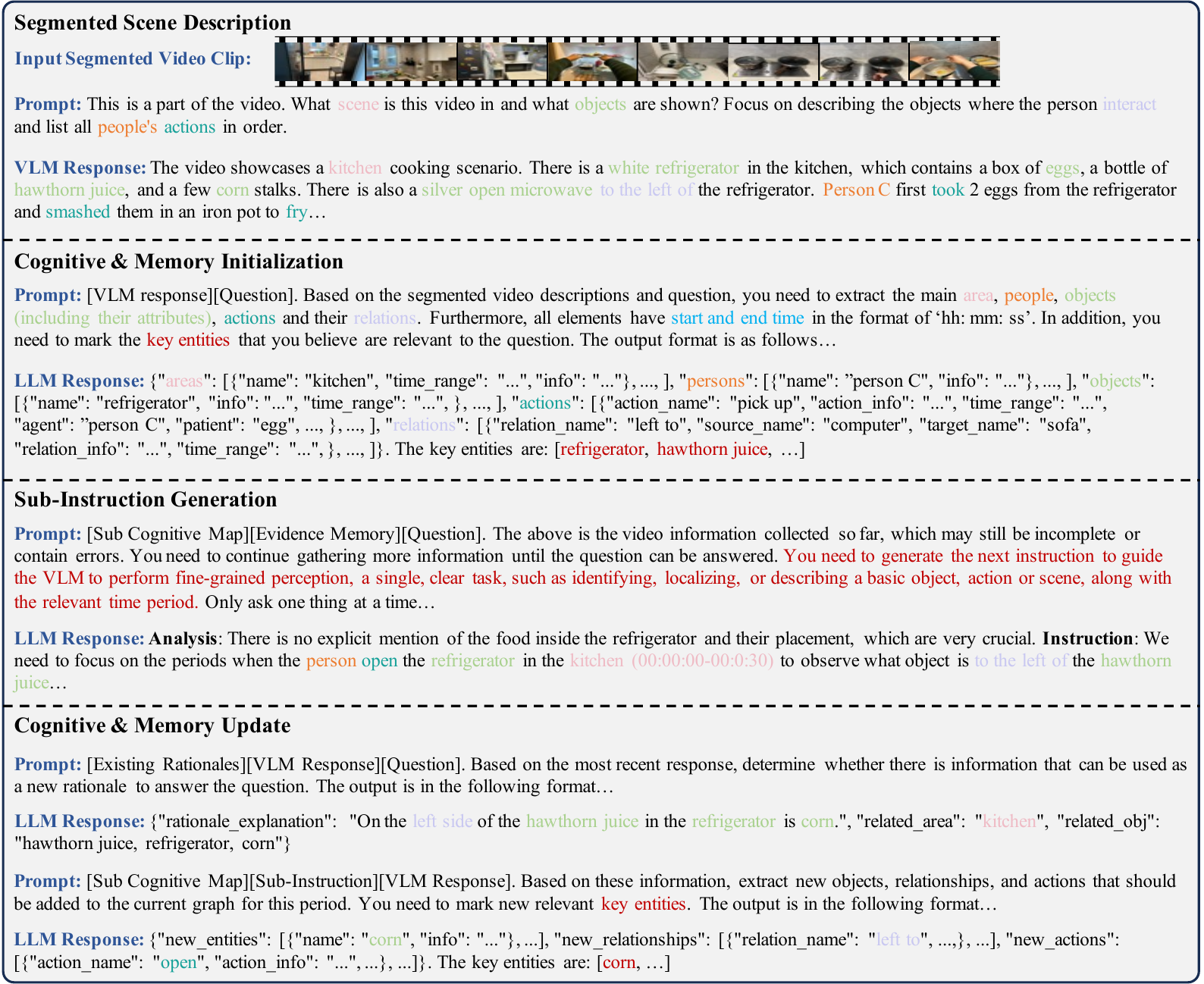}
    \caption{\textbf{Prompt for LLM-VLM Synergy.} For brevity, the prompt here is abbreviated. Please refer to appendix for the complete version.}
    \label{fig:prompt}
\end{figure*}

\noindent
\textbf{Cognitive Map \& Evidence Memory Update.}
Importantly, the cognitive map $\mathcal{M}$ is not a static structure. As stated in Stage (1) of Sec. \ref{ssec:framework}, the initial $\mathcal{M}^{(0)}$ is constructed by parsing VLM-generated scene descriptions with the LLM. Subsequently, it is iteratively refined through interactions between the LLM and VLM. Specifically, in reasoning iterations, the LLM leverages the instruction $I$ and the previous state memory to decompose the task into a set of subtasks $T=\left \{ T_1,T_2,\dots ,T_n \right \}$, guiding the VLM to perceive the corresponding video segments $V_{T_i}$ to extract task-relevant visual evidence to update the cognitive map. Formally, the update at iteration $i$ is defined as:
\begin{equation}
    \mathcal{M}^{(i)} = Update(\mathcal{M}^{(i-1)}, \text{VLM}(V_{T_i}, T_i) ),
\end{equation}

where $Update(\cdot)$ is a function that integrates new information while ensuring consistency. Specifically, it first extracts the relevant temporal subgraph from $\mathcal{M}^{(i-1)}$ as context. Then, a specialized LLM-guided prompt instructs the LLM to identify new entities, relations, or actions from the VLM output that are absent in the existing graph. This process also handles conflict resolution by following a temporal precedence principle: newer observations from the VLM override older, contradictory information, prompting the LLM to update or remove outdated elements. All updates (additions, deletions, modifications) are processed atomically to maintain integrity. Finally, key entities relevant to the question are managed to ensure the map remains focused and does not become overloaded with irrelevant information. This enables the cognitive map to evolve progressively, offering structured and up-to-date grounding for final answer reasoning. Similarly, evidence memory is also constantly updated in reasoning iterations:
\begin{equation}
    \mathcal{E}^{(i)} = Update(\mathcal{E}^{(i-1)}, \text{VLM}(V_{T_i}, T_i) ).
\end{equation}

% The memory is updated incrementally as subtasks progress:
% \begin{equation}
%     \mathcal{E}^{(i)} = \mathcal{E}^{(i-1)}\cup \left \{ a_i \right \}.
% \end{equation}

\noindent
\textbf{Integrated Reasoning \& Answer Generation.}
The cognitive map and evidence memory are integrated to form the complete reasoning context, upon which the LLM generates the response for the current step:
\begin{equation}
    R_i = \text{LLM}(\mathcal{M}^{(i)}, \mathcal{E}^{(i)}, I).
\end{equation}
If the LLM determines that reasoning is complete or the maximum number of iterations has been reached, the output will be returned as the final response $R$. Otherwise, the output will be regarded as the next subtask $T_{i+1}$ to guide the VLM to continue updating the $\mathcal{M}$ and $\mathcal{E}$:
\begin{equation}
    R_i=\begin{cases}
  R& \text{ if } Exit=True, \\
  T_{i+1}& \text{ if } Exit=False.
\end{cases}
\end{equation}

\subsection{Prompt for Linguistic-Visual Synergy‌}
\label{ssec:synergy}

In the CLiViS framework, the LLM-VLM collaboration is orchestrated by meticulously designed prompts aligning high-level reasoning with visual perception.
As illustrated in Figure \ref{fig:prompt}, these prompts guide the key stages of the pipeline: (a) \textbf{Segmented Scene Description}, performed by the VLM to generate coarse-grained visual captions; (b) \textbf{Cognitive and Memory Initialization}, where the LLM processes scene descriptions to extract entities, actions, and relations, and identifies key entities; (c) \textbf{Sub-Instruction Generation}, where the LLM analyzes existing information and formulates targeted instructions for subsequent VLM perception; (d) \textbf{Cognitive Map and Evidence Memory Update}, in which the LLM integrates the VLM outputs to refine the current graph and memory. Further details on the prompt design and its impact on task-specific reasoning can be found in the supplementary materials.
\section{Experiments}

\subsection{Experimental Setup}
\label{ssec:exp}

\begin{table*}
\centering
\caption{\textbf{Performance comparison with other methods across different benchmarks.} All results are reproduced under unified settings with the same model configurations (\textit{e.g.}, FPS, temperature). Best results are marked in \textbf{bold}, and the second-best is \underline{underlined}.}
\vspace{-6pt}
\scalebox{0.88}{
\begin{tabular}{c|ccc ccc c|c}
\hline
\multirow{2}{*}{Models} 
& \multicolumn{3}{c}{OpenEQA} 
& \multicolumn{3}{c}{EgoTempo} 
& \multirow{2}{*}{EgoSchema} 
& \multirow{2}{*}{Avg.} \\ 
\cmidrule(lr){2-4} \cmidrule(lr){5-7}
& $<30$ s & $\ge 30$ s & All 
& $<30$ s & $\ge 30$ s & All 
&  &  \\ 
\hline
\rowcolor[HTML]{EFEFEF} 
\multicolumn{9}{l}{\cellcolor[HTML]{EFEFEF}{\color[HTML]{000000} Socratic-based Models}} \\ \hline
\multicolumn{1}{c|}{Qwen2.5-VL + Qwen2.5-Max} & 31.9 & 21.9 & 23.0 & 6.7 & 5.3 & 5.8  & \multicolumn{1}{c|}{58.6} & 29.1 \\
\multicolumn{1}{c|}{Qwen2.5-VL + DeepSeek-V3} & 32.8 & 19.9 & 23.5 & 6.2 & 4.6 & 5.2  & \multicolumn{1}{c|}{55.6} & 28.1 \\
\multicolumn{1}{c|}{InternVL3 + Qwen2.5-Max} & 7.8 & 7.5 & 7.5  & 3.6 & 1.6 & 2.4 & \multicolumn{1}{c|}{57.2} & 22.4 \\
\multicolumn{1}{c|}{InternVL3 + DeepSeek-V3} & 12.1 & 9.8 & 10.0 & 2.6 & 3.3 & 3.0  & \multicolumn{1}{c|}{54.0} & 22.3 \\
\multicolumn{1}{c|}{VideoLLaMA3 + Qwen2.5-Max} & 16.4 & 8.3 & 9.2  & 5.1 & 4.6 & 4.8 & \multicolumn{1}{c|}{61.8} & 25.3 \\
\multicolumn{1}{c|}{VideoLLaMA3 + DeepSeek-V3} & 19.0 & 8.2 & 9.4  & 8.2 & 5.3 & 6.4  & \multicolumn{1}{c|}{55.2} & 23.7 \\ \hline
\rowcolor[HTML]{EFEFEF} 
\multicolumn{9}{l}{\cellcolor[HTML]{EFEFEF}End-to-End VLM Models} \\ \hline
\multicolumn{1}{c|}{Qwen2-VL} & 37.1 & 29.9 & 30.7 & 10.3 & 10.5 & 9.6 & \multicolumn{1}{c|}{48.4} & 29.6\\
\multicolumn{1}{c|}{Qwen2.5-VL} & 49.1 & 39.7 & 40.7 & 18.9 & 14.5 & 16.2 & \multicolumn{1}{c|}{64.8} & 40.6 \\
\multicolumn{1}{c|}{InternVL2.5} & 46.6 & 35.3 & 36.5 & 9.7 & 12.1 & 10.2 & \multicolumn{1}{c|}{63.2} & 36.6 \\
\multicolumn{1}{c|}{InternVL3} & 50.9 & 53.8 & 53.6 & 15.8 & 17.7 & 17.0 & \multicolumn{1}{c|}{66.6} & 45.7 \\
\multicolumn{1}{c|}{VideoLLaMA3} & \underline{57.8} & \underline{57.0} & \underline{57.1} & 21.4 & 18.7 & 19.8 & \multicolumn{1}{c|}{62.2} & 46.4 \\ \hline
\rowcolor[HTML]{EFEFEF} 
\multicolumn{9}{l}{\cellcolor[HTML]{EFEFEF}Video Reasoning Models} \\ \hline
\multicolumn{1}{c|}{VideoAgent} & 4.3 & 8.3 & 7.9  & 8.2 & 7.2 & 7.6  & \multicolumn{1}{c|}{62.0} & 25.8 \\
\multicolumn{1}{c|}{VideoTree} & 18.9 & 15.5& 16.4    & 18.9 & 13.9 & 14.8 & \multicolumn{1}{c|}{60.0} & 30.4 \\
\multicolumn{1}{c|}{LVNet} & - & -& -    & - & - & - & \multicolumn{1}{c|}{58.8} & - \\
\multicolumn{1}{c|}{AlanaVLM} & 42.2 & 34.9& 35.7    & 17.9 & 15.4 & 16.4 & \multicolumn{1}{c|}{32.2} & 28.1 \\
\multicolumn{1}{c|}{Video-R1} & 52.6 & 40.6 & 41.9 & 18.0 & 15.4 & 16.4 & \multicolumn{1}{c|}{46.6} & 35.0 \\
\rowcolor[HTML]{DAE8FC} 
\multicolumn{1}{c|}{\cellcolor[HTML]{DAE8FC}CLiViS (Qwen2.5-VL)} & 52.6 & 46.2 & 46.9 & 21.4 & 17.8 & 19.6 & \multicolumn{1}{c|}{\cellcolor[HTML]{DAE8FC} \underline{68.2}} & 44.9 \\
\rowcolor[HTML]{DAE8FC} 
\multicolumn{1}{c|}{\cellcolor[HTML]{DAE8FC}CLiViS (InternVL3)} & 51.7 & 55.9 & 55.4 & \textbf{28.1} & \underline{19.4} & \underline{23.0} & \multicolumn{1}{c|}{\cellcolor[HTML]{DAE8FC} \textbf{69.4} } & \textbf{49.3} \\
\rowcolor[HTML]{DAE8FC} 
\multicolumn{1}{c|}{\cellcolor[HTML]{DAE8FC}CLiViS (VideoLLaMA3)} & \textbf{58.6} & \textbf{57.2} & \textbf{57.3} & \underline{23.5} & \textbf{23.4} & \textbf{23.4} & \multicolumn{1}{c|}{\cellcolor[HTML]{DAE8FC} 64.8} & \underline{48.4} \\ \hline
\end{tabular}}
\label{tab:main}
\end{table*}

\noindent
\textbf{Benchmarks.}
We conduct experiments on three real-world egocentric video question answering benchmarks: 
(1) \textbf{OpenEQA} \cite{majumdar2024openeqa} combines ‌over 1,600 QA pairs‌ from real-world environments. We evaluate on a subset of 1,079 QA pairs with corresponding videos from ScanNet.
(2) \textbf{EgoSchema} \cite{mangalam2023egoschema}, built on Ego4D\cite{grauman2022ego4d}, consists of 3-minutes egocentric clips depicting daily activities. We apply the official validation set of 500 questions.
(3) \textbf{EgoTempo} \cite{plizzari2025omnia} is a recently-proposed challenging benchmark for evaluating temporal reasoning. It includes 500 QA cases across 10 categories, enabling fine-grained assessment of dynamic visual reasoning in long-term egocentric videos.
% Please refer to Appendix for additional details.

\noindent
\textbf{Evaluation Metrics.}
We adopt the standard \textit{accuracy} metric to evaluate all methods. For the multiple-choice benchmark (EgoSchema), we report overall accuracy on 3-minute videos.
For the open-ended benchmarks (OpenEQA and EgoTempo), we follow their original protocols and utilize Qwen2.5-Max to score model responses against reference answers on a 5-point Likert scale, with scores $\geq 4$ considered correct. 
To assess the impact of video length, we further split the videos into short ($<30$ sec) and long ($\geq 30$ sec) groups.

\noindent
\textbf{Compared Methods.}
We compare the proposed CLiViS with three paradigms of methods: 
(1) \textbf{Socratic-based models}: two-stage pipelines combining VLMs (Qwen2.5-VL \cite{bai2025qwen25vl}, InternVL3 \cite{zhu2025internvl3}, and VideoLLaMA3 \cite{zhang2025videollama}) with LLMs (Qwen2.5-Max \cite{yang2024qwen25} and DeepSeek-V3 \cite{liu2024deepseek});
(2) \textbf{End-to-end VLMs}: Open-source end-to-end VLMs including Qwen2-VL \cite{wang2024qwen2vl}, Qwen2.5-VL, InternVL2.5 \cite{chen2024expanding}, InternVL3, and VideoLLaMA3.
(3) \textbf{Video reasoning models}: VideoAgent \cite{fan2024videoagent}, VideoTree \cite{wang2024videotree}, LVNet \cite{Park2024TooMF}, and Video-R1 \cite{feng2025video}. The former three are also training-free methods, and we adopt Qwen2.5-Max as their LLM and replace GPT-4o with Qwen2.5-VL in LVNet for fair comparison. Notably, VideoTree and LVNet are designed for multiple‑choice setting, therefore their performance on open-ended benchmarks are omitted.

\begin{table*}
    \centering
    \caption{\textbf{Model-agnostic effectiveness.} CLiViS achieves consistent improvements across different VLMs, showing strong model-agnostic effectiveness. All experiments use Qwen2.5-Max as the LLM.}
    % All experiments use Qwen2.5-Max as the language model.
    \vspace{-6pt}
    \scalebox{0.88}{
    \begin{tabular}{cc|ccc|c}
\hline
\multicolumn{2}{c|}{Models} & OpenEQA & EgoTempo & EgoSchema & Avg. \\ \hline
& baseline & 40.7 & 16.2 & 64.8 & 40.6 \\
\multirow{-2}{*}{Qwen2.5-VL} & +CLiViS & \cellcolor[HTML]{DAE8FC}{46.9 (\textcolor[HTML]{009901}{+6.2})} & \cellcolor[HTML]{DAE8FC} {19.6 (\textcolor[HTML]{009901}{+3.4})} & \cellcolor[HTML]{DAE8FC} {68.2 (\textcolor[HTML]{009901}{+3.4})} & \cellcolor[HTML]{DAE8FC} {44.9 (\textcolor[HTML]{009901}{+4.3})}\\ \hline
& baseline & 53.6 & 17.0 & 66.6 & 45.7 \\
\multirow{-2}{*}{InternVL3} & +CLiViS  & \cellcolor[HTML]{DAE8FC} {55.4 (\textcolor[HTML]{009901}{+1.8})} & \cellcolor[HTML]{DAE8FC} {23.0 (\textcolor[HTML]{009901}{+6.0})} & \cellcolor[HTML]{DAE8FC} {69.4 (\textcolor[HTML]{009901}{+2.8})} & \cellcolor[HTML]{DAE8FC} {49.3 (\textcolor[HTML]{009901}{+3.6})} \\ \hline
& baseline & 57.1 & 19.8 & 62.2 & 46.4 \\
\multirow{-2}{*}{VideoLLaMA3} & +CLiViS & \cellcolor[HTML]{DAE8FC} {57.3 (\textcolor[HTML]{009901}{+0.2})}& \cellcolor[HTML]{DAE8FC} {23.4 (\textcolor[HTML]{009901}{+3.6})} & \cellcolor[HTML]{DAE8FC} {64.8 (\textcolor[HTML]{009901}{+2.6})} & \cellcolor[HTML]{DAE8FC} {48.4 (\textcolor[HTML]{009901}{+2.0})}\\ \hline
\end{tabular}}
    \label{tab:model}
\end{table*}

\subsection{Implementation Details}
\label{ssec:parameter}
Unless specified, all experiments use Qwen2.5-Max as the LLM. All models (\textit{e.g.}, Qwen2.5-VL, InternVL2.5, and VideoLLaMA3) are 7B–8B variants. To facilitate long-term video reasoning, we segment videos into 30-second intervals, leaving shorter clips unsegmented. The maximum number of dialogue rounds is limited to 10. See the appendix for more experimental details.

\subsection{Main Results}
\noindent
\textbf{Consistent State‑of‑the‑Art Performance.} Table \ref{tab:main} reports the results on three representative benchmarks. Across all three benchmarks, CLiViS achieves the state-of-the-art performance ({55.4\% on OpenEQA, 23.0\% on EgoTempo, and 69.4\% on EgoSchema}), yielding an overall average accuracy of 49.4\%. This consistent performance across diverse benchmarks demonstrate its ability to reconcile perception and reasoning in a unified, training‑free framework for embodied visual reasoning.

\noindent
\textbf{Advantage Across Paradigms.} CLiViS outperforms the strongest model within each paradigm, giving a 20.2\% gain over Socratic-based models, a 2.9\% improvement over end-to-end VLMs, and a 14.3\% increase compared to video reasoning methods. This improvement arises from CLiViS’s integration of three complementary paradigms, combining the language reasoning capabilities of Socratic-based strategy, the open‑vocabulary perception of end‑to‑end VLMs, and the multi‑step inference of video reasoning methods, within a unified, training‑free framework. This iterative approach is key to outperforming other training-free methods like VideoTree/LVNet. Their "perceive-then-reason" pipeline relies on a static one-step LLM reasoning. This rigid process struggles with complex verification, leading to lower EgoSchema performance, and is ill-suited for the open-ended reasoning required by OpenEQA and EgoTempo. In contrast, CLiViS's LLM "planner" actively guides VLM perception in a hypothesis-verification loop, dynamically updating its Cognitive Map. This iterative synergy allows the system to resolve ambiguities and verify complex relationships, fundamentally overcoming the limitations of single-step reasoning.

\noindent
\textbf{Impact of Video Length.} The margin of improvement achieved by CLiViS over baseline models increases with video duration. Using Qwen2.5‑VL on OpenEQA as an example, CLiViS outperforms the baseline by 3.5\% on videos less than 30s and by 6.5\% on videos over 30s. This demonstrates our framework’s superior ability to aggregate long‑range visual and semantic cues via the iterative LLM–VLM collaboration and dynamic cognitive map.

\subsection{Model-agnostic Effectiveness}
Table \ref{tab:model} shows that CLiViS consistently improves performance across three different VLM backbones when paired with the same LLM (Qwen2.5‑Max). On EgoTempo, CLiViS increases the accuracy of InternVL3 from 17.0\% to 23.0\% (+6.0\%), while also improving Qwen2.5‑VL and VideoLLaMA3 by 3.4\% and 3.6\%, respectively. Similar trends are observed on OpenEQA and EgoSchema. On average, CLiViS achieves a 4.3\% improvement with Qwen2.5‑VL across all three datasets. Notably, even when base model VideoLLaMA3 already performs strongly on OpenEQA (57.1\%), CLiViS further enhances its accuracy by {0.2\%}. These results validate the model-agnostic effectiveness of our framework, demonstrating its consistent ability to enhance reasoning performance regardless of the specific VLM backbone.

\begin{figure}
  \centering
  \begin{minipage}[t]{0.45\textwidth}
    % \vspace{0pt}
    \centering
    \captionof{table}{\textbf{Ablation studies on EgoTempo.} We evaluate each component of CLiViS using the combination of InternVL3 and Qwen2.5-Max.}
    \label{tab:ablation}
    % \vspace{8pt}
    \scalebox{1.0}{
\begin{tabular}{l|c}
\hline
Variants            & Accuracy    \\ \hline
baseline (VLM only)    & 17.0 (\textcolor[HTML]{CB0000}{-6.0}) \\
w/o Navigation Graph & 20.6 (\textcolor[HTML]{CB0000}{-2.4}) \\
w/o Relation Graph & 21.4 (\textcolor[HTML]{CB0000}{-1.6}) \\
w/o Evidence Memory & 22.4 (\textcolor[HTML]{CB0000}{-0.6}) \\
w/o multi-round interaction & 12.5 (\textcolor[HTML]{CB0000}{-10.5})\\
w/ VLM for reasoning & 10.6 (\textcolor[HTML]{CB0000}{-12.4})\\ \hline
full model (VLM + LLM) & 23.0        \\ \hline
\end{tabular}}
  \end{minipage}
  \hfill
  \vspace{0pt}
\end{figure}

\subsection{Ablation Studies}
Table \ref{tab:ablation} presents the ablation studies on EgoTempo. Removing the \textbf{Navigation Graph} causes a 2.4\% drop, confirming its critical role in temporal localization. Removing the \textbf{Relation Graph} results in a 1.6\% drop, validating its importance for fine-grained spatial context. Disabling the \textbf{Evidence Memory} further reduces performance by 0.6\%, confirming its utility in tracking rationales. Notably, the most significant degradation occurs from altering the core framework components. Collapsing the multi-round LLM-VLM interaction into a single round causes a severe 10.5\% drop, highlighting that iterative linguistic-visual synergy is critical. Even more striking, replacing the LLM (Qwen2.5-Max) with a VLM (InternVL3) for the high-level reasoning task results in a massive 12.4\% performance drop. This result confirms that a powerful, dedicated LLM is not merely supplementary but fundamentally necessary for complex embodied reasoning.

\subsection{Inference Latency Analysis}
\label{ssec:latency}

\begin{table}
\centering
\caption{\textbf{Latency and Accuracy on EgoSchema.}}
\label{tab:latency}
% \vspace{8pt}
\scalebox{0.9}{
\begin{tabular}{l|c c}
\hline
Method & Latency (s) & Accuracy (\%) \\ \hline
VideoTree & \textbf{71} & 60.0 \\
VideoAgent & 644 & 62.0 \\
\rowcolor[HTML]{DAE8FC} % This highlights your method
CLiViS (Ours) & 195 & \textbf{69.4} \\ \hline
\end{tabular}
}

\end{table}

\noindent
To evaluate the accuracy-computation trade-off, we conducted a runtime analysis on the EgoSchema benchmark (InternVL3 model), presented in Table \ref{tab:latency}. While real-world latency remains a challenge for current zero-shot methods, the results show CLiViS offers a superior balance. Although VideoTree is faster, our method surpasses its accuracy by a significant margin (+9.4\%). Conversely, CLiViS is substantially faster than VideoAgent while also achieving stronger performance.
\section{Conclusion}
In this paper, we introduce CLiViS, a novel paradigm of embodied visual reasoning that harnesses the complementary strengths of LLMs and VLMs. Unlike prior approaches that rely on static captions or end-to-end models with limited reasoning capability, CLiViS establishes a collaborative framework: the LLM acts as a high-level planner, while the VLM serves as a perceptual executor guided by task-specific prompts. To bridge perception and reasoning, we introduce a dynamic cognitive map and an evidence memory, which together support structured, multi-step inference. Extensive experiments across diverse benchmarks confirm the effectiveness and generality of our approach.
\section*{Acknowledgements}
\label{sec:ack}
This work was supported by Shanghai Municipal Science and Technology Major Project (No. 2025SHZDZX025G16), National Key R\&D Program of China (No. 2025ZD1801501), NSFC grant (No. 62477014 and No. 62136002), Ministry of Education Research Joint Fund Project (No. 8091B042239), Shanghai Knowledge Service Platform Project (No. ZF1213), and Shanghai Trusted Industry Internet Software Collaborative Innovation Center.
{
    \small
    \bibliographystyle{ieeenat_fullname}
    \bibliography{main}
}

\clearpage
\setcounter{page}{1}
\maketitlesupplementary

In this supplementary material, we provide additional details and analyses to complement the main paper. 
Sec.~\ref{appendix:config} elaborates on the implementation details, covering experimental parameters, relation graph construction, and compute resources. 
Sec.~\ref{Appendix:supplementry_Results} presents extended quantitative results with a fine-grained evaluation on the EgoTempo benchmark.
Sec.~\ref{sec:frontier_comparison} provides a comparative analysis with frontier multimodal models regarding accuracy.
This is followed by an analysis of reasoning dynamics in Sec.~\ref{sec:reasoning}. 
Sec.~\ref{Appendix:qualitative_results} visualizes the model's decision-making process through comprehensive qualitative examples. 
All the prompts used in our framework are placed in the \texttt{prompts\_txt} folder in the supplementary material.
% \section{Technical Appendices and Supplementary Material}
% Technical appendices with additional results, figures, graphs and proofs may be submitted with the paper submission before the full submission deadline (see above), or as a separate PDF in the ZIP file below before the supplementary material deadline. There is no page limit for the technical appendices.

\section{More Implementation Details}
\label{appendix:config}
\paragraph{Additional Experimental Parameters.}
During evaluation, we set the temperature to $0.5$ and $\text{top\_p}=0.9$ for LLM components, while VLMs use a temperature of $0.3$ and $\text{top\_p}=0.9$. Considering hardware limitations and efficiency, we preprocess videos by sampling frames at 0.5 FPS on OpenEQA and EgoSchema, while limiting frame count to 32 for EgoTempo. We also pre-scale all videos from the EgoTempo and OpenEQA datasets to a resolution of 480p prior to experimentation.

 \paragraph{Implementation Details of Relation Graph.}
The relational graph is implemented using Neo4j, a graph database system. Neo4j is an open source NoSQL graph database designed to efficiently store and process highly interconnected data, supporting complex graph traversals and relationship queries. It is widely used in applications such as knowledge graphs. As input to the LLM during inference, the subgraph extracted from the full relational graph includes the following components: (1) all designated key entities (nodes); (2) all paths between key nodes, along with intermediate nodes and relationships on those paths; (3) nodes and relationships directly connected to each key node; (4) all associated nodes and relationships linked to activity-type nodes. In addition, we limit the maximum path length to 10 steps when retrieving paths between key nodes to ensure the query efficiency.

\paragraph{Experiments Compute Resources.}
All VLMs used in our experiments, including Qwen2-VL (7B), Qwen2.5-VL (7B), InternVL2.5 (8B), InternVL3 (8B), and VideoLLaMA3 (7B), are deployed and executed locally. The hardware setup for experiments consists of 2 A100 80G GPUs and 4 A6000 48G GPUs. LLMs such as Qwen2.5-Max and DeepSeek-V3 are accessed via APIs provided by a cloud computing platform.

\begin{table*}[ht]
\centering
\caption{\textbf{Performance Comparison on Various Question Categories of the EgoTempo benchmark.} Best results are marked in \textbf{bold}, and the second-best is \underline{underlined}.}
\vspace{-6pt}
\scalebox{0.8}{
\begin{tabular}{c|cccccccccc|c}
\hline
\multicolumn{1}{c|}{Models} 
& AO & LO & SR & OS & AS & OA & TE & FU & AC & OC & Avg. \\
\hline
\rowcolor[HTML]{EFEFEF} 
\multicolumn{12}{l}{\cellcolor[HTML]{EFEFEF}{\color[HTML]{000000} Socratic-based Models}} \\ \hline
\multicolumn{1}{c|}{Qwen2.5-VL + Qwen2.5-Max} & 8.0 & 10.0 & 0.0 & 4.0 & 0.0 & 6.0 & 6.0 & 18.0 & 0.0 & 6.0 & 5.8 \\
\multicolumn{1}{c|}{Qwen2.5-VL + DeepSeek-V3} & 10.0 & 4.0 & 4.0 & 2.0 & 4.0 & 12.0 & 4.0 & 10.0 & 0.0 & 2.0 & 5.2 \\
\multicolumn{1}{c|}{InternVL3 + Qwen2.5-Max} & 6.0 & 2.0 & 2.0 & 4.0 & 0.0 & 4.0 & 2.0 & 4.0 & 0.0 & 0.0 & 2.4 \\
\multicolumn{1}{c|}{InternVL3 + DeepSeek-V3} & 10.0 & 2.0 & 0.0 & 0.0 & 0.0 & 4.0 & 4.0 & 8.0 & 0.0 & 2.0 & 3.0 \\
\multicolumn{1}{c|}{VideoLLaMA3 + Qwen2.5-Max} & 10.0 & 4.0 & 2.0 & 2.0 & 2.0 & 6.0 & 10.0 & 10.0 & 0.0 & 2.0 & 4.8 \\
\multicolumn{1}{c|}{VideoLLaMA3 + DeepSeek-V3} & 14.0 & 2.0 & 0.0 & 2.0 & 4.0 & 8.0 & 10.0 & 18.0 & 4.0 & 2.0 & 6.4 \\
\hline
\rowcolor[HTML]{EFEFEF} 
\multicolumn{12}{l}{\cellcolor[HTML]{EFEFEF}End-to-End VLM Models} \\ \hline
\multicolumn{1}{c|}{Qwen2-VL} & 18.0 & 8.0 & \underline{28.0} & 4.0 & 0.0 & 6.0 & 10.0 & 2.0 & 18.0 & 10.0 & 9.6 \\
\multicolumn{1}{c|}{Qwen2.5-VL} & \underline{28.0} & 18.0 & 24.0 & 10.0 & 4.0 & 6.0 & 12.0 & \textbf{26.0} & 16.0 & 18.0 & 16.2 \\
\multicolumn{1}{c|}{InternVL2.5} & 20.0 & 14.0 & 14.0 & 4.0 & 2.0 & 8.0 & 10.0 & 4.0 & 12.0 & 24.0 & 10.2 \\
\multicolumn{1}{c|}{InternVL3} & 22.0 & 22.0 & 22.0 & 6.0 & 2.0 & 6.0 & 20.0 & 8.0 & 20.0 & \textbf{42.0} & 17.0 \\
\multicolumn{1}{c|}{VideoLLaMA3} & \underline{28.0} & \underline{26.0} & 24.0 & 4.0 & 2.0 & \textbf{24.0} & \underline{22.0} & 12.0 & 14.0 & \textbf{42.0} & 19.8 \\
\hline
\rowcolor[HTML]{EFEFEF} 
\multicolumn{12}{l}{\cellcolor[HTML]{EFEFEF}Video Reasoning Models} \\ 
\hline
\multicolumn{1}{c|}{VideoAgent} & 20.0 & 10.0 & 0.0 & 6.0 & 2.0 & 6.0 & 12.0 & 14.0 & 0.0 & 6.0 & 7.6 \\
\multicolumn{1}{c|}{Video-R1} & 18.0 & 12.0 & 22.0 & 4.0 & 0.0 & 12.0 & 14.0 & 18.0 & \textbf{28.0} & \underline{36.0} & 16.4 \\
\rowcolor[HTML]{DAE8FC} 
\multicolumn{1}{c|}{\cellcolor[HTML]{DAE8FC}CLiViS (Qwen2.5-VL)} & 20.0 & 18.0 & \textbf{32.0} & 12.0 & \underline{12.0} & 18.0 & 16.0 & \underline{22.0} & \underline{26.0} & 16.0 & 19.6 \\
\rowcolor[HTML]{DAE8FC} 
\multicolumn{1}{c|}{\cellcolor[HTML]{DAE8FC}CLiViS (InternVL3)} & \textbf{36.0} & \textbf{30.0} & 22.0 & \textbf{28.0} & \textbf{16.0} & 12.0 & \textbf{28.0} & 16.0 & 8.0 & 32.0 & \underline{23.0} \\
\rowcolor[HTML]{DAE8FC} 
\multicolumn{1}{c|}{\cellcolor[HTML]{DAE8FC}CLiViS (VideoLLaMA3)} & \textbf{36.0} & 22.0 & \textbf{32.0} & \underline{16.0} & \textbf{16.0} & \underline{22.0} & 20.0 & \underline{22.0} & 18.0 & 30.0 & \textbf{23.4} \\ 
\hline
\end{tabular}}
\label{tab:categories}
\end{table*}

\section{Supplementary Results}
\label{Appendix:supplementry_Results}

To further understand how different methods handle diverse instructions, we conduct a fine-grained evaluation on the EgoTempo dataset. The questions in EgoTempo are categorizes into 10 distinct types, including Action-Specific Objects (\textbf{AO}), Locating Objects (\textbf{LO}), Spatial Relations (\textbf{SR}), Object and Action Sequences (\textbf{OS}, \textbf{AS}), Object-Specific Actions (\textbf{OA}), Temporal Events (\textbf{TE}), Future Prediction (\textbf{FU}), Action and Object Counting (\textbf{AC}, \textbf{OC}).

As shown in Table \ref{tab:categories}, CLiViS consistently outperforms all compared methods across most categories. Notably, compared to others, CLiViS delivers substantial improvements in reasoning‐intensive categories such as Spatial Relations (SR), Action Sequences (AS), and Temporal Event (TE), demonstrating its superior capability for multi‑step, long‑horizon visual reasoning.

% Compared to the strongest end-to-end VLM baseline (VideoLLaMA3, 19.8%), CLiViS yields clear gains in complex reasoning categories such as Spatial Relations (SR), Object Sequence (OS), and Future Prediction (FU), demonstrating its superior ability to handle multi-step, instruction-driven visual reasoning.

\begin{table}[ht]
\centering
\caption{\textbf{Comparison with Frontier Multimodal Models on EgoTempo.}}
\label{tab:frontier_comparison}
\vspace{-6pt}
\scalebox{0.9}{
\begin{tabular}{l|c}
\toprule
Models & Accuracy (\%) \\
\midrule
GPT-4.1 & 34.2 \\
Gemini-2.5-flash & 27.4 \\
\midrule
CLiViS (w/ GPT-4.1) & \textbf{36.8} \\
CLiViS (w/ Gemini-2.5-flash) & \textbf{30.8} \\
\bottomrule
\end{tabular}}
\end{table}

\section{Comparison with Frontier Multimodal Models}
\label{sec:frontier_comparison}

We have conducted additional experiments using GPT-4.1 and Gemini-2.5-flash on the EgoTempo benchmark to evaluate performance. The results are presented in Table~\ref{tab:frontier_comparison}.

As shown in the table, CLiViS delivers superior accuracy compared to standalone models (36.8\% vs 34.2\% for GPT-based systems, 30.8\% vs 27.4\% for Gemini-based systems). This performance gap highlights CLiViS's effectiveness in complex embodied reasoning tasks, where its iterative LLM-VLM interaction and dynamic cognitive mapping enable deeper analysis compared to standard end-to-end approaches.

\begin{figure}[ht]
    \centering
    \includegraphics[width=0.8\linewidth]{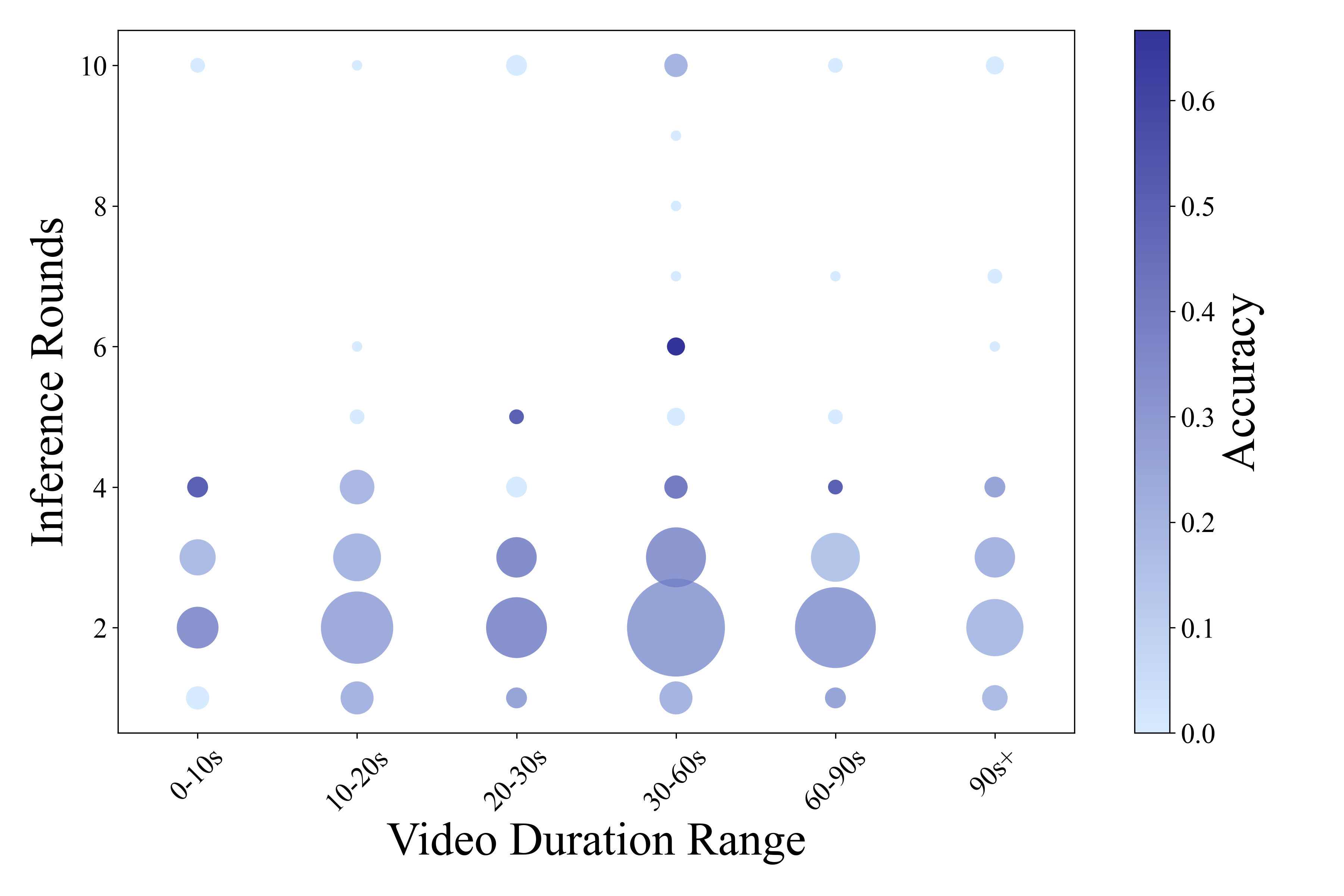}
    \caption{\textbf{Correlation between reasoning rounds and video duration.}}
    \label{Fig:reasoning}
\end{figure}

\section{Reasoning Dynamics}
\label{sec:reasoning}
Figure \ref{Fig:reasoning} illustrates a multivariate bubble chart, where the horizontal axis represents video duration, the vertical axis indicates the number of inference rounds performed by CLiViS, bubble color represents the accuracy (darker is better) and bubble size reflects the number of samples. From the visualization, we have the following observations: (1) While most videos can be resolved within 2–3 reasoning rounds, the distribution of reasoning steps gradually shifts toward 3–5 rounds as video duration increases, indicating that CLiViS adaptively allocates deeper reasoning for complex inputs. (2) Within each duration range, a positive correlation emerges between the reasoning rounds and the accuracy, revealing that deeper reasoning may lead to higher-quality responses. These observations highlight CLiViS’s ability to dynamically modulate reasoning rounds based on temporal complexity, achieving a favorable balance between efficiency and performance.

\section{More Qualitative Results}
\label{Appendix:qualitative_results}

\begin{figure*}[p]
    \centering
    \includegraphics[width=0.85\textwidth]{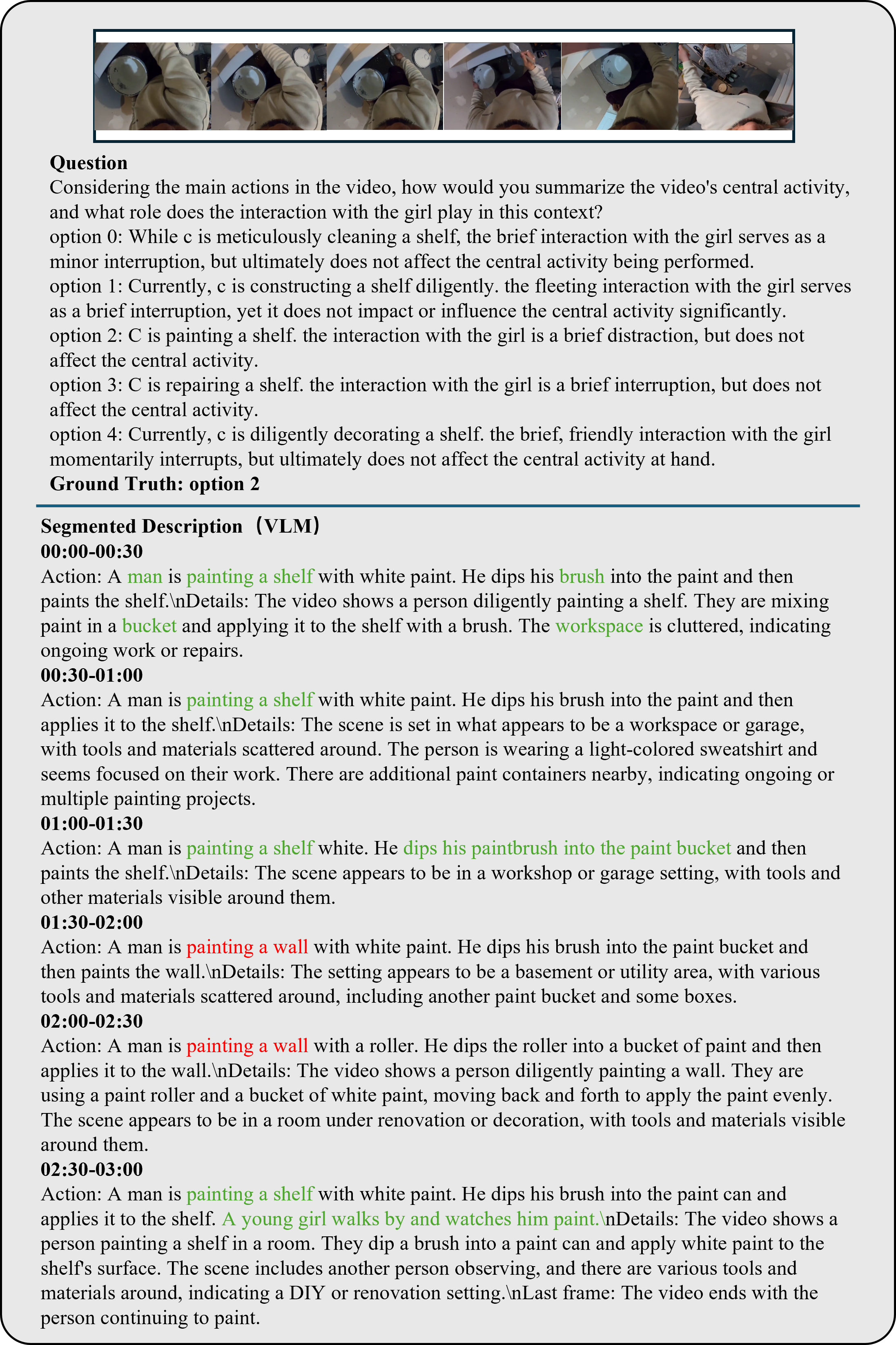}
    \caption{Qualitative results of Segmented Description.}
    \label{fig:example_1}
\end{figure*}

\begin{figure*}[p]
    \centering
    \includegraphics[width=0.9\textwidth]{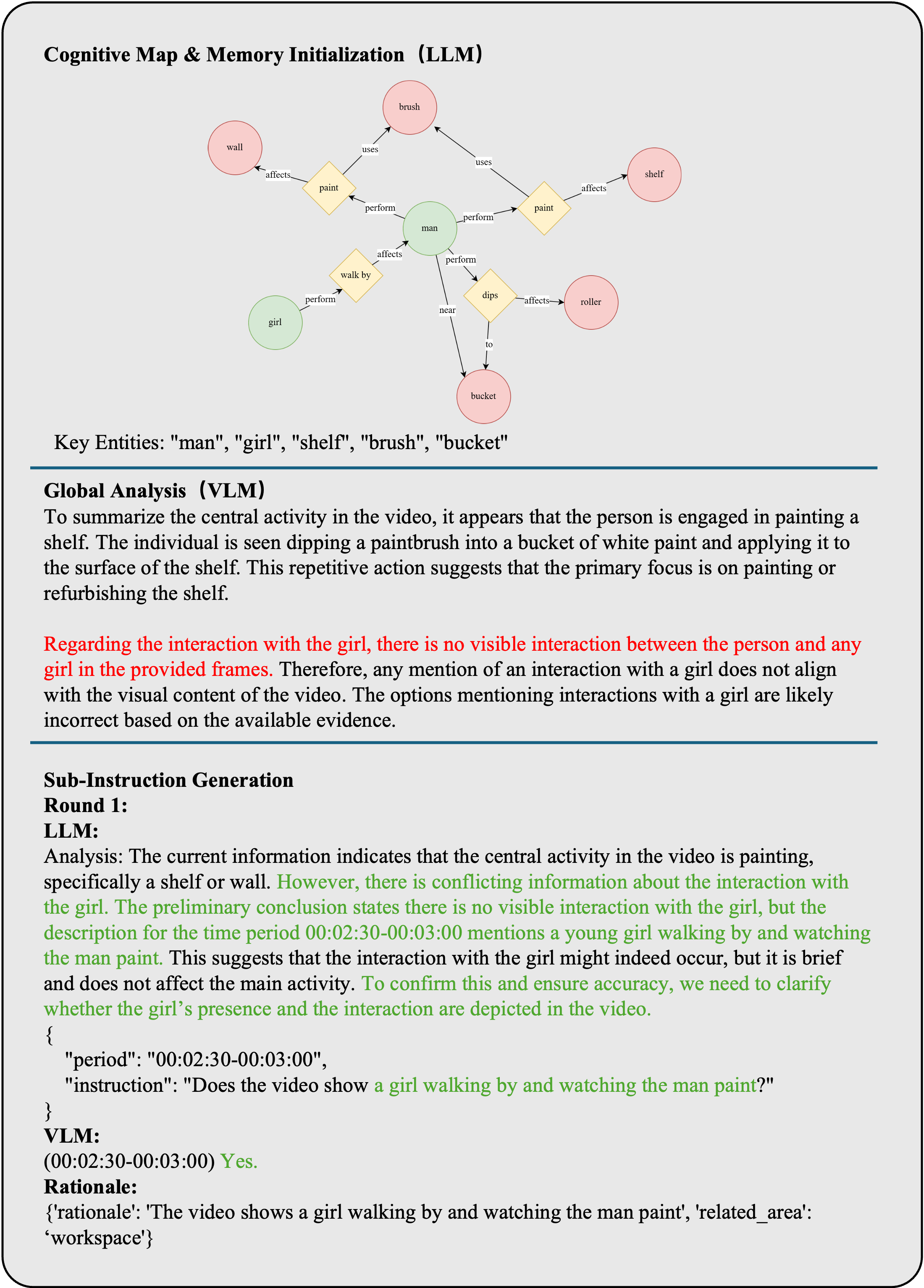}
    \caption{Qualitative results of Cognitive Map and Sub-Instrction Generation.}
    \label{fig:example_2}
\end{figure*}

\begin{figure*}[p]
    \centering
    \includegraphics[width=0.9\textwidth]{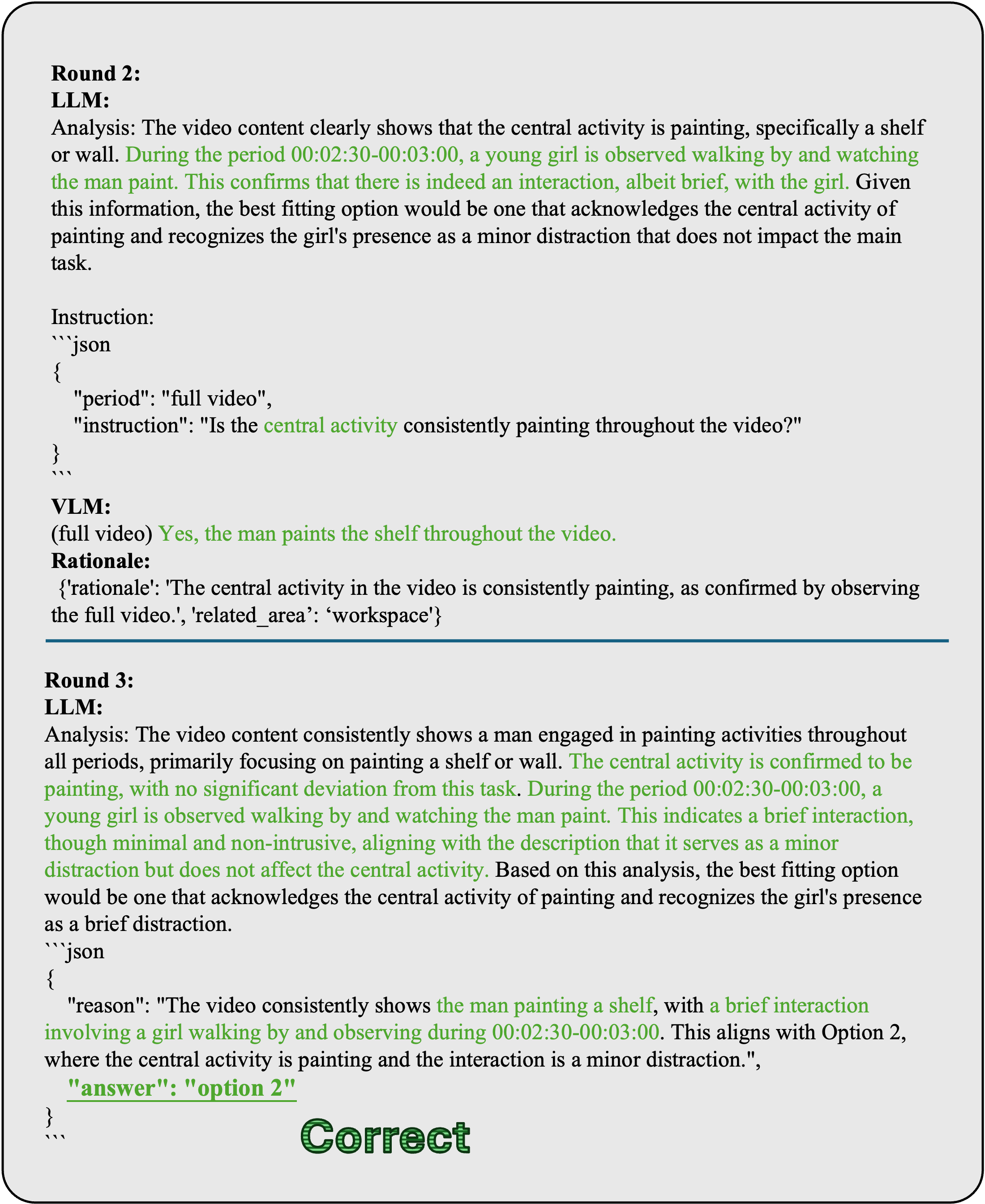}
    \caption{Qualitative results of Iterative Reasoning and Answer Generation.}
    \label{fig:example_3}
\end{figure*}

To illustrate CLiViS’s reasoning process and interpretability, we present a qualitative case from the EgoSchema dataset in Figure \ref{fig:example_1} to Figure \ref{fig:example_3}, that traces every step from cognitive initialization to sub-instruction generation and execution and cognitive update. We highlight relevant information throughout the reasoning trace, marking correct details in green and erroneous ones in red. Overall, this example validates four key advantages of CLiViS:
\begin{itemize}
    \item \textbf{Comprehensive scene parsing.} Segmenting the video into clips and generating per-segment descriptions enable the VLM to better capture fine-grained details, resulting in a more comprehensive and precise cognitive graph.
    \item \textbf{Clear task decomposition.} At each reasoning iteration, the LLM allocates the next perception task based on accumulated cognition, ensuring that every subtask yields fresh and relevant insights.
    \item \textbf{Self-correcting ability.} It demonstrates the ability to proactively detect and resolve errors or inconsistencies in the VLM-generated descriptions, significantly improving the overall robustness and logical consistency of the reasoning.
    \item \textbf{Strong interpretability.} Every subtask, VLM response, and cognitive map update is fully traceable, providing transparent decision paths.
\end{itemize}

These results demonstrate that CLiViS not only maintains high accuracy in embodied visual reasoning tasks involving long‑range dependencies and complex semantic instructions, but also delivers an auditable modular inference chain, validating both its effectiveness and interpretability.

% \newpage

\newpage

\newpage

\clearpage

\FloatBarrier

\end{document}